\documentclass{article}

\usepackage{times}
\usepackage{graphicx}
\usepackage{multirow}
\usepackage{natbib}

\usepackage{algorithm}
\usepackage{caption}
\usepackage{subcaption}
\usepackage{algorithmic}
\usepackage{amsmath}
\usepackage{mathtools}
\usepackage{amsfonts}
\usepackage{booktabs}
\usepackage{graphicx}
\usepackage{amssymb}
\usepackage{soul}
\usepackage{color}

\def \ie {\emph{i.e.}}

\usepackage{hyperref}

\usepackage[accepted]{icml2017}

\icmltitlerunning{Unsupervised Learning by Predicting Noise}

\begin{document}

\twocolumn[
\icmltitle{Unsupervised Learning by Predicting Noise}


\begin{icmlauthorlist}
\icmlauthor{Piotr Bojanowski}{fair}
\icmlauthor{Armand Joulin}{fair}
\end{icmlauthorlist}
\icmlaffiliation{fair}{Facebook AI Research}

\icmlcorrespondingauthor{Piotr Bojanowski}{bojanowski@fb.com}

\icmlkeywords{Unsupervised Learning, Convolutional Neural Networks, Discriminative Clustering}

\vskip 0.3in
]

\printAffiliationsAndNotice{}  

\begin{abstract}
Convolutional neural networks provide visual features that perform remarkably well in many computer vision applications.
However, training these networks requires significant amounts of supervision.
This paper introduces a generic framework to train deep networks, end-to-end, with no supervision.
We propose to fix a set of target representations, called \emph{Noise As Targets} (NAT), and to constrain the deep features to align to them.
This domain agnostic approach avoids the standard unsupervised learning issues of trivial solutions and collapsing of features.
Thanks to a stochastic batch reassignment strategy and a separable square loss function, it scales to millions of images.
The proposed approach produces representations that perform on par with state-of-the-art unsupervised methods on ImageNet and \textsc{Pascal} VOC.
\end{abstract}

\section{Introduction}

In recent years, convolutional neural networks, or \emph{convnets}~\citep{fukushima1982neocognitron,LBDHHHJ89} have
pushed the limits of computer vision~\citep{KSH12, he16}, leading to important progress
in a variety of tasks, like object detection~\citep{girshick2015fast} or image
segmentation~\citep{pinheiro2015learning}.  
Key to this success is their ability to produce features that easily transfer
to new domains when trained on massive databases of labeled
images~\citep{razavian14,oquab2014learning} or weakly-supervised data~\cite{joulin16}. 
However, human annotations may introduce unforeseen
bias that could limit the potential of learned features to capture subtle information
hidden in a vast collection of images.  

Several strategies exist to learn deep convolutional
features with no annotation~\citep{DKD16}.  They either try to
capture a signal from the source as a form of \emph{self-supervision}~\citep{DGE15,WG15}
or learn the underlying distribution of 
images~\citep{vincent2010stacked,goodfellow2014generative}.
While some of these approaches obtain promising performance in
transfer learning~\citep{DKD16,WG15}, they do not explicitly aim to learn
discriminative features.
Some attempts were made with retrieval based approaches~\citep{DSRB14} and clustering~\citep{YPB16,LSZU16},
but they are hard to scale and have only been tested on small datasets.
Unfortunately, as in the supervised case, a lot of data is required to learn good representations.

In this work, we propose a novel discriminative framework designed to learn deep architectures on massive amounts of data. 
Our approach is general, but we focus on convnets since they require millions of images to produce good features.  
Similar to self-organizing maps~\cite{kohonen1982self, martinetz1991neural},
we map deep features to a set of predefined representations in a low dimensional space.
As opposed to these approaches, we aim to learn the features in a end-to-end fashion, 
which traditionally suffers from a feature collapsing problem.  Our approach
deals with this issue by fixing the target representations and aligning them to our features.
These representations are sampled from a uninformative distribution and we use this \emph{Noise As Targets} (NAT).
Our approach also shares some similarities with standard clustering approches like $k$-means~\citep{lloyd1982least}
or discriminative clustering~\cite{Bach07}.

In addition, we propose an online algorithm able to scale to massive image databases like ImageNet~\citep{deng2009imagenet}. 
Importantly, our approach is barely less efficient to train than 
standard supervised approaches and can re-use any 
optimization procedure designed for them.
This is achieved by using a quadratic loss as in~\cite{CSTTW17} and a fast approximation of the Hungarian algorithm.
We show the potential of our approach by training end-to-end on ImageNet a standard architecture, namely AlexNet~\citep{KSH12} with no supervision. 

We test the quality of our features on
several image classification problems, following the setting of~\citet{DKD16}.
We are on par with state-of-the-art unsupervised and self-supervised learning approaches while being much simpler to train
and to scale.

The paper is organized as follows: after a brief review of
the related work in Section~\ref{sec:related}, we present
our approach in Section~\ref{sec:method}. We then 
validate our solution with several experiments and 
comparisons with standard unsupervised and self-supervised approaches in Section~\ref{sec:exp}.


\section{Related work}
\label{sec:related}

Several approaches have been recently proposed to tackle the problem of deep unsupervised learning~\cite{CN12, MKHS14, DSRB14}.
Some of them are based on a clustering loss~\cite{XGF16,YPB16,LSZU16}, but they are not tested at a scale comparable to that of supervised convnet training.
\citet{CN12} uses $k$-means to pre-train convnets, by learning each layer sequentially in a bottom-up fashion. 
In our work, we train the convnet end-to-end with a loss that shares similarities with $k$-means.
Closer to our work, \citet{DSRB14} proposes to train convnets by solving a retrieval problem.
They assign a class per image and its transformation. 
In contrast to our work, this approach can hardly scale to more than a few hundred of thousands of images, and requires a custom-tailored architecture while we use a standard AlexNet.

Another traditional approach for learning visual representations in an unsupervised manner is to define a parametrized mapping between a predefined random variable and a set of images.  
Traditional examples of this approach are variational autoencoders~\cite{kingma2013auto}, generative adversarial networks~\citep{goodfellow2014generative}, and to a lesser extent, noisy autoencoders~\citep{vincent2010stacked}.  
In our work, we are doing the opposite; that is, we map images to a predefined random variable.
This allows us to re-use standard convolutional networks and greatly simplifies the training.

\paragraph{Generative adversarial networks.} Among those approaches, generative adversarial
networks (GANs)~\cite{goodfellow2014generative,denton2015deep,DKD16} share another
similarity with our approach, namely
they are explicitly minimizing a discriminative loss 
to learn their features. 
While these models cannot learn an inverse mapping, \citet{DKD16} recently
proposed to add an encoder to extract visual features
from GANs. 
Like ours, their encoder can be any standard convolutional network.
However, their loss aims at differentiating real and generated images, while
we are aiming directly at differentiating between images. This makes our
approach much simpler and faster to train, since we do not need to learn
the generator nor the discriminator.

\paragraph{Self-supervision.}
Recently, a lot of work has explored \emph{self-supervison}:  
leveraging supervision contained in the input signal~\cite{DGE15,NF16, pathak2016context}.
In the same vein as word2vec~\cite{mikolov2013efficient}, ~\citet{DGE15} show
that spatial context is a strong signal to learn visual features. \citet{NF16}
have further extended this work.  Others have shown that temporal coherence in videos also provides a
signal that can be used to learn  powerful visual
features~\cite{ACM15,JG15,WG15}. In particular,~\citet{WG15} show that such
features provide promising performance on ImageNet.
In contrast to our work, these approaches are domain dependent since they
require explicit derivation of weak supervision directly from the input.

\paragraph{Autoencoders.}
Many have also used autoencoders with a reconstruction
loss~\citep{bengio2007greedy,huang2007unsupervised,masci2011stacked}. The idea
is to encode and decode an image, while minimizing the loss between the decoded
and original images. Once trained, the encoder produces image features and
the decoder can be used to generate images from codes.  The decoder is often a
fully connected network~\cite{huang2007unsupervised} or a deconvolutional
network~\citep{masci2011stacked, whatwhere} but can be more sophisticated, like
a PixelCNN network~\citep{van2016conditional}.

\paragraph{Self-organizing map.} This family of unsupervised methods aims at
learning a low dimensional representation of the data that preserves certain topological properties~\citep{kohonen1982self, vesanto2000clustering}.  In
particular, Neural Gas~\citep{martinetz1991neural} aligns feature vectors to
the input data. Each input datum is then assigned to one of these vectors in a
winner-takes-all manner. These feature vectors are in spirit similar to our
target representations and we use a similar assignment strategy. In contrast to our work,
the target vectors are not fixed and aligned to the input vectors. Since
we primarly aim at learning the input features, we do the opposite. 

\paragraph{Discriminative clustering.}
Many methods have been proposed to use discriminative losses for
clustering~\cite{xu2004maximum,Bach07,krause2010discriminative,JouBacICML12}.
In particular,
\citet{Bach07} shows that the ridge regression loss could be use to
learn discriminative clusters. It has been successfully
applied to several computer vision applications, like
object discovery~\citep{Joulin10,TJLF14} or
video/text alignment~\cite{Bojanowski_ICCV13,bojanowski2014weakly,ramanathan2014linking}. 
In this work, we show that a similar framework can be
designed for neural networks. As opposed to \citet{xu2004maximum},
we address the empty assignment problems by restricting the 
set of possible reassignments to permutations rather than
using global linear constrains the assignments.
Our assignments can be updated online, allowing our approach to scale to very large datasets.


\section{Method}
\label{sec:method}

\begin{figure}[t]
\centering
\includegraphics[width=\linewidth]{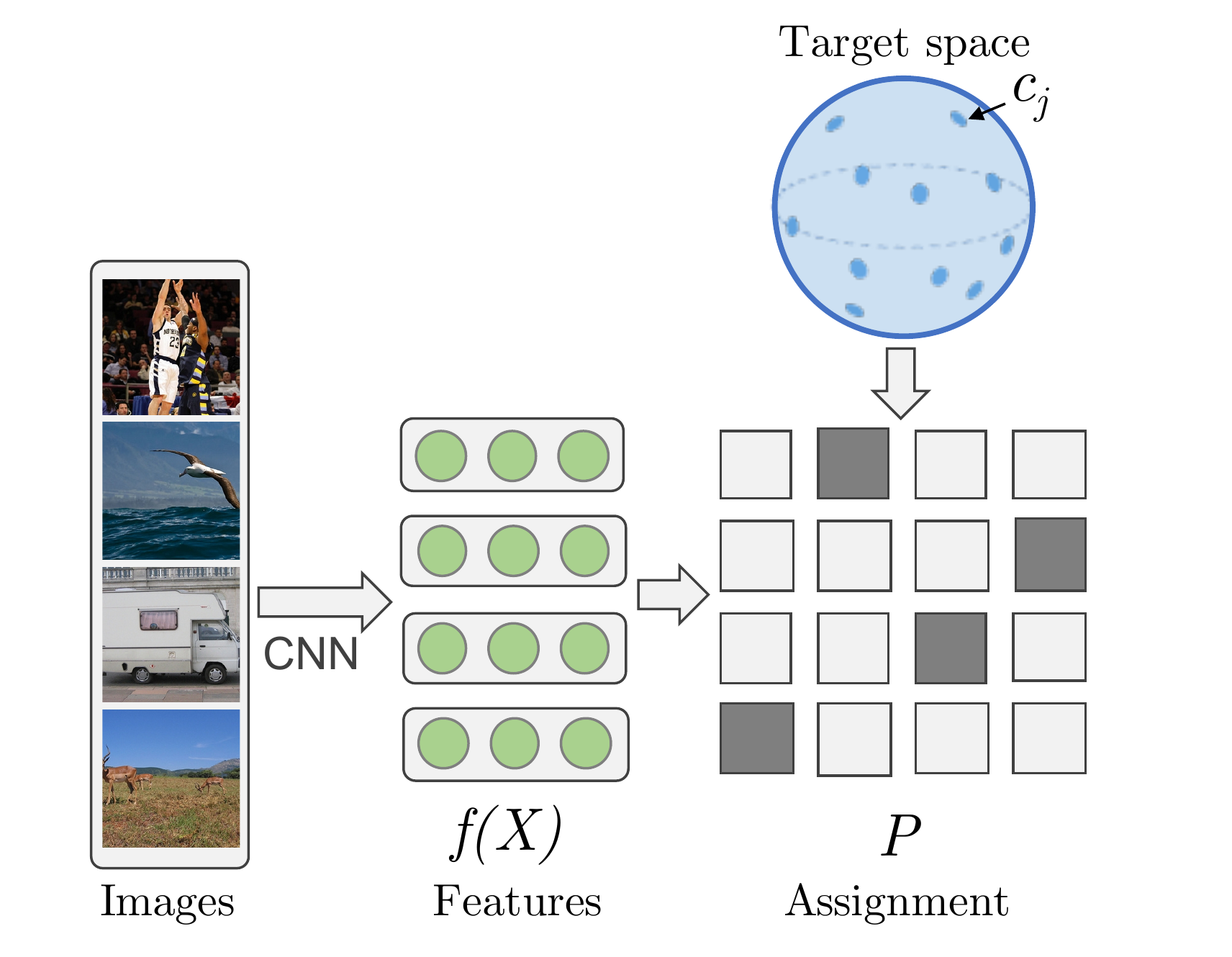}
\caption{
  Our approach takes a set of images, computes their deep features with a convolutional network and matches them to
  a set of predefined targets from a low dimensional space. The parameters of the network are learned
  by aligning the features to the targets.
}
\label{fig:pullfig}
\end{figure}

In this section, we present our model and discuss its relations with several clustering approaches including $k$-means.
Figure~\ref{fig:pullfig} shows an overview of our approach.
We also show that it can be trained on massive datasets using an online procedure.
Finally, we provide all the implementation details.

\subsection{Unsupervised learning}
\label{sec:unsup}
We are interested in learning visual features with no supervision.
These features are produced by applying a parametrized mapping $f_\theta$ to the images.
In the presence of supervision, the parameters $\theta$ are learned by minimizing a loss function between the
features produced by this mapping and some given targets, e.g., labels.
In absence of supervision, there is no clear target representations and we thus need to learn them as well.
More precisely, given a set of $n$ images $x_i$, we jointly learn the parameters $\theta$ of the mapping $f_\theta$,
and some target vectors $y_i$:
\begin{equation}
  \label{eq:pb}
  \min_{\theta} \ \frac{1}{n} \sum_{i=1}^n \ \min_{y_i \in \mathbb{R}^d} \ \ell(f_\theta(x_i), y_i),
\end{equation}
where $d$ is the dimension of target vectors.
In the rest of the paper, we use matrix notations, \ie, we denote by $Y$
the matrix whose rows are the target representations $y_i$,
and by $X$ the matrix whose rows are the images $x_i$.
With a slight abuse of notation, we denote by $f_\theta(X)$ the $n \times d$ matrix of features whose rows are obtained by applying the function $f_\theta$ to each image independently.

\paragraph{Choosing the loss function.}
In the supervised setting, a popular choice for the loss $\ell$ is the softmax function.
However, computing this loss is linear in the number of targets, making it impractical
for large output spaces~\citep{goodman2001classes}.
While there are workarounds to scale these losses to large output spaces,
\citet{CSTTW17} has recently shown that using a squared $\ell_2$ distance works well in many supervised settings, as long as the final activations are unit normalized.
This loss only requires access to a single target per sample, making its computation independent of the number of targets.
This leads to the following problem:
\begin{equation}\label{eq:mat}
  \min_{\theta} \ \min_{Y\in\mathbb{R}^{n\times d}} \ \frac{1}{2n} \|f_\theta(X) - Y\|_F^2,
\end{equation}
where we still denote by $f_\theta(X)$ the unit normalized features.

\paragraph{Using fixed target representations.}
Directly solving the problem defined in Eq.~(\ref{eq:mat}) would lead to a
representation collapsing problem: all the images would be assigned to the same representation~\citep{xu2004maximum}.
We avoid this issue by fixing a set of $k$ predefined target representations and matching them
to the visual features.
More precisely, the matrix $Y$ is defined as the product of a matrix $C$ containing these $k$ representations and an assignment matrix $P$ in $\{0,1\}^{n \times k}$, \ie,
\begin{equation}
  Y = P C.
\end{equation}
Note that we can assume that $k$ is greater than $n$ with no loss of generality (by duplicating representations otherwise).
Each image is assigned to a different target and each target can only be assigned once.
This leads to a set $\mathcal{P}$ of constraints for the assignment matrices:
\begin{equation}\label{eq:p}
\mathcal{P} = \{ P \in \{0,1\}^{n\times k} ~ | ~ P1_k \le 1_n, ~ P^\top 1_n = 1_k\}.
\end{equation}
This formulation forces the visual features to be diversified, avoiding the collapsing issue at the cost of fixing the target representations.
Predefining these targets is an issue if their number $k$ is small, which is why we are interested in the case
where $k$ is at least as large as the number $n$ of images.

\paragraph{Choosing the target representations.}
Until now, we have not discussed the set of target representations stored in $C$.
A simple choice for the targets would be to take $k$ elements of the canonical basis of $\mathbb{R}^d$.
If $d$ is larger than $n$, this formulation would be similar to the framework of~\citet{DSRB14}, and is
impractical for large $n$.
On the other hand, if $d$ is smaller than $n$, this formulation is equivalent to the discriminative
clustering approach of~\citet{Bach07}.
Choosing such targets makes very strong assumptions on the nature of the underlying problem.
Indeed, it assumes that each image belongs to a unique class and that all classes are orthogonal.
While this assumption might be true for some classification datasets, it does not generalize to huge image collections
nor capture subtle similarities between images belonging to different classes.

Since our features are unit normalized,
another natural choice is to uniformly sample target vectors on the $\ell_2$ unit sphere.
Note that the dimension $d$ will then directly influence the level of correlation
between representations, \ie, the correlation is inversely proportional to the square root of $d$.
Using this \emph{Noise As Targets} (NAT), Eq.~(\ref{eq:mat}) is now equivalent to:
\begin{equation}\label{eq:lin}
\max_\theta \max_{P\in\mathcal{P}} \text{Tr} \left  ( P C f_\theta(X)^\top \right ).
\end{equation}
This problem can be interpreted as mapping deep features to a uniform distribution over a manifold,
namely the $d$-dimension $\ell_2$ sphere.
Using $k$ predefined representations is a discrete approximation of this manifold that justifies
the restriction of the mapping matrices to the set $\mathcal{P}$ of $1$-to-$1$ assignment matrices.
In some sense, we are optimizing a crude approximation of the earth mover's distance between the distribution
of deep features and a given target distribution~\cite{rubner1998metric}.

\paragraph{Relation to clustering approaches.}
Using the same notations as in Eq.~(\ref{eq:lin}),
several clustering approaches share similarities with our method.
In the linear case, spherical $k$-means minimizes the same loss
function w.r.t. $P$ and $C$, \ie,
\begin{equation*}
\max_C\max_{P\in\mathcal{Q}} \text{tr}\left(PCX^T\right).
\end{equation*}
A key difference is the set $\mathcal{Q}$ of assignment matrices:
\begin{equation*}
  \mathcal{Q} = \{P\in\{0,1\}^{n\times k}~|~P 1_k = 1_n\}.
\end{equation*}
This set only guarantees that each data point is assigned to a single target representation.
Once we jointly learn the features and the assignment, this set does not prevent
the collapsing of the data points to a single target representation.

Another similar clustering approach is Diffrac~\citep{Bach07}. Their loss is equivalent to ours
in the case of unit normalized features.
Their set $\mathcal{R}$ of assignment matrices, however, is different:
\begin{equation*}
\mathcal{R} = \{P\in\{0,1\}^{n\times k}~|~P^\top 1_n \ge c 1_k\},
\end{equation*}
where $c>0$ is some fixed parameter.
While restricting the assignment matrices to this set prevents the collapsing issue,
it introduces global constraints that are not suited for online optimization.
This makes their approach hard to scale to large datasets.


\begin{algorithm}[t]
  \caption{
    Stochastic optimization of Eq.~(\ref{eq:lin}).
  }
  \label{alg1}
  \begin{algorithmic}
    \REQUIRE $T$ batches of images, $\lambda_0>0$
    \FOR{$t = \{1,\dots, T\}$}
    \item
      Obtain batch $b$ and representations $r$
    \item
      Compute $f_\theta(X_b)$
    \item
      Compute $P^*$ by minimizing Eq. (\ref{eq:mat}) w.r.t. $P$
    \item
      Compute $\nabla_\theta L(\theta)$ from Eq.~(\ref{eq:mat}) with $P^*$
    \item
      Update $\theta\leftarrow \theta - \lambda_t \nabla_\theta L(\theta)$
    \ENDFOR
  \end{algorithmic}
\end{algorithm}

\subsection{Optimization}
\label{sec:opt}
In this section, we describe how to efficiently optimize the cost function described in Eq.~(\ref{eq:lin}).
In particular, we explore approximated updates of the assignment matrix that are compatible
with online optimization schemes, like stochastic gradient descent (SGD).

\paragraph{Updating the assignment matrix $P$.}
Directly solving for the optimal assignment requires to evaluate the distances between all the $n$ features and the $k$ representations.
In order to efficiently solve this problem, we first reduce the number $k$ of representations to $n$.
This limits the set $\mathcal{P}$ to the set of permutation matrices, \ie,
\begin{equation}\label{eq:newp}
  \mathcal{P} = \{ P \in \{0,1\}^{n\times n} ~ | ~ P1_n = 1_n, ~ P^\top 1_n = 1_n\}.
\end{equation}
Restricting the problem defined in Eq.~(\ref{eq:lin}) to this set, the linear assignment problem in $P$ can be solved exactly
with the Hungarian algorithm~\citep{kuhn1955hungarian}, but at the prohibitive cost of $O(n^3)$.

Instead, we perform stochastic updates of the matrix.
Given a batch of samples, we optimize the assignment matrix $P$ on its restriction to this batch.
Given a subset $\mathcal{B}$ of $b$ distinct images,
we only update the $b\times b$ square sub matrix $P_\mathcal{B}$ obtained by restricting $P$
to these $b$ images and their corresponding targets.
In other words, each image can only be re-assigned to a target that was previously assigned to another image in the batch.
This procedure has a complexity of $O(b^3)$ per batch, leading to an overall complexity of $O(n b^2)$,
which is linear in the number of data points.
We perform this update before updating the parameters $\theta$ of our features, in an on-line manner.
Note that this simple procedure would not have been possible if $k > n$;
we would have had to also consider the $k-n$ unassigned representations.

\paragraph{Stochastic gradient descent.}
Apart from the update of the assignment matrix $P$,
we use the same optimization scheme as standard supervised approaches, \ie,
SGD with batch normalization~\citep{ioffe2015}.
As noted by~\citet{CSTTW17}, batch normalization plays a crucial role
when optimizing the $\l_2$ loss, as it avoids exploding gradients.
For each batch $b$ of images, we first perform a forward pass to compute the distance between the images and the corresponding subset of target representations $r$.
The Hungarian algorithm is then used on these distances to obtain the optimal reassignments within the batch.
Once the assignments are updated, we use the chain rule in order to compute the gradients of all our parameters.
Our optimization algorithm is summarized in Algorithm~\ref{alg1}.


\subsection{Implementation details}
Our experiments solely focus on learning visual features with convnets.
All the details required to train these architectures with our approach are described below.
Most of them are standard tricks used in the usual supervised setting.

\paragraph{Deep features.}
To ensure a fair empirical comparison with previous work, we follow~\citet{WG15} and use an AlexNet architecture.
We train it end to end using our unsupervised loss function.
We subsequently test the quality of the learned visual feature by re-training a classifier on top.
During transfer learning, we consider the output of the last convolutional layer as our features as in~\citet{razavian14}.
We use the same multi-layer perceptron (MLP) as in~\citet{KSH12} for the classifier.

\paragraph{Pre-processing.}
We observe in practice that pre-processing the images greatly helps the quality of our learned features.
As in~\citet{huang2007unsupervised}, we use image gradients instead of the images to avoid trivial solutions like clustering according to colors.
Using this preprocessing is not surprising since most hand-made features like SIFT or HoG are based on image gradients~\citep{L99,dalal2005histograms}.
In addition to this pre-processing, we also perform all the standard image transformations that are commonly applied in the supervised setting~\cite{KSH12},
such as random cropping and flipping of images.

\begin{table}[t]
  \centering
  \begin{tabular}{@{}cccc@{}}
    \toprule
     && Softmax & Square loss  \\
    \midrule
    ImageNet && 59.2 & 58.4 \\
    \bottomrule
  \end{tabular}
  \caption{
    Comparison between the softmax and the square loss for supervised object classification on ImageNet.
    The architecture is an AlexNet. The features are unit normalized for the square loss~\cite{CSTTW17}. We report the accuracy on the validation set.
  }
  \label{tab:l2loss}
\end{table}

\paragraph{Optimization details.}
We project the output of the network on the $\ell_2$ sphere as in~\citet{CSTTW17}.
The network is trained with SGD with a batch size of $256$.
During the first $t_0$ batches, we use a constant step size.
After $t_0$ batches, we use a linear decay of the step size, \ie, $l_t = \frac{l_0}{1 + \gamma [t - t_0]_+}$.
Unless mentioned otherwise, we permute the assignments within batches every $3$ epochs.
For the transfer learning experiments, we follow the guideline described in~\citet{DKD16}.

\section{Experiments}
\label{sec:exp}
We perform several experiments to validate different design choices in NAT.
We then evaluate the quality of our features by comparing them to state-of-the-art unsupervised approaches on several auxiliary supervised tasks, namely object classification on ImageNet and object classification and detection of \textsc{Pascal} VOC 2007~\citep{EVWWZ10}.

\paragraph{Transfering the features.}
In order to measure the quality of our features, we measure their performance on transfer learning.
We freeze the parameters of all the convolutional layers and overwrite the parameters of the MLP classifier with random Gaussian weights.
We precisely follow the training and testing procedure that is specific to each of the datasets following~\citet{DKD16}.

\paragraph{Datasets and baselines.}
We use the training set of ImageNet to learn our convolutional network~\citep{deng2009imagenet}.
This dataset is composed of $1,281,167$ images that belong to $1,000$ object categories.
For the transfer learning experiments, we also consider \textsc{Pascal} VOC 2007.
In addition to fully supervised approaches~\citep{KSH12},
we compare our method to several unsupervised approaches, \ie, autoencoder, GAN and BiGAN as reported in~\citet{DKD16}.
We also compare to self-supervised approaches, \ie, \citet{ACM15,DGE15,pathak2016context,WG15} and \citet{zhang2016colorful}.
Finally we compare to state-of-the-art hand-made features, \ie, SIFT with Fisher Vectors (SIFT+FV)~\citep{akata2014good}. They reduce the Fisher Vectors to a $4,096$ dimensional vector with PCA, and apply an $8,192$ unit $3$-layer MLP on top.


\subsection{Detailed analysis}
In this section, we validate some of our design choices, like the loss function, representations and the influences of some parameters on the quality of our features.
All the experiments are run on ImageNet.

\paragraph{Softmax \emph{versus} square loss.}
Table~\ref{tab:l2loss} compares the performance of an AlexNet trained with a softmax and a square loss.
We report the accuracy on the validation set.
The square loss requires the features to be unit normalized to avoid exploding gradients.
As previously observed by~\citet{CSTTW17}, the performances are similar, hence validating our choice of loss function.

\paragraph{Effect of image preprocessing.}
In supervised classification, image pre-processing is not frequently used, and transformations that remove information are usually avoided.
In the unsupervised case, however, we observe that is it is preferable to work with simpler inputs as it avoids learning trivial features.
In particular, we observe that using grayscale image gradients greatly helps our method, as mentioned in Sec.~\ref{sec:method}.
In order to verify that this preprocessing does not destroy crucial information, we propose to evaluate its effect on supervised classification.
We also compare with high-pass filtering.
Table~\ref{tab:degradation} shows the impact of this preprocessing methods on the accuracy of an AlexNet on the validation set of ImageNet.
None of these pre-processings degrade the perform significantly, meaning that the information related to gradients are sufficient for
object classification. This experiment confirms that such pre-processing does not lead to a significant drop in the
upper bound performance for our model.

\begin{table}[t]
  \centering
  \begin{tabular}{@{}rrrrr@{}}
    \toprule
    & clean & high-pass & sobel \\
    \midrule
    acc@1 & 59.7 & 58.5 & 57.4 \\
    \bottomrule
  \end{tabular}
  \caption{
    Performance of supervised models with various image pre-processings applied.
    We train an AlexNet on ImageNet, and report classification accuracy.
  }
  \label{tab:degradation}
\end{table}

\paragraph{Continuous \emph{versus} discrete representations.}
We compare our choice for the target vectors to those commonly used for clustering, \ie, elements of the canonical basis of a $k$ dimensional space.
Such discrete representation make a strong assumption on the underlying structure of the problem, that it can be linearly separated in $k$ different classes.
This assumption holds for ImageNet giving a fair advantage to this discrete representation. We test this representation with k in $\{10^3, 10^4, 10^5\}$, which
is a range well-suited for the $1,000$ classes of ImageNet.
The matrix $C$ contains $n/k$ replications of $k$ elements of the canonical basis. This assumes that the clusters are balanced, which is verified on ImageNet.

We compare these cluster-like representations to our continuous target vectors on the transfer task on ImageNet.
Using discrete targets achieves an accuracy of $19\%$, which is significantly worse that our best performance, \ie, $33.5\%$.
A possible explanation is that binary vectors induce sharp discontinuous distances between representations.
Such distances are hard to optimize over and may result in early convergence to poorer local minima.

\begin{figure}[t]
  \centering
  \includegraphics[height=10em]{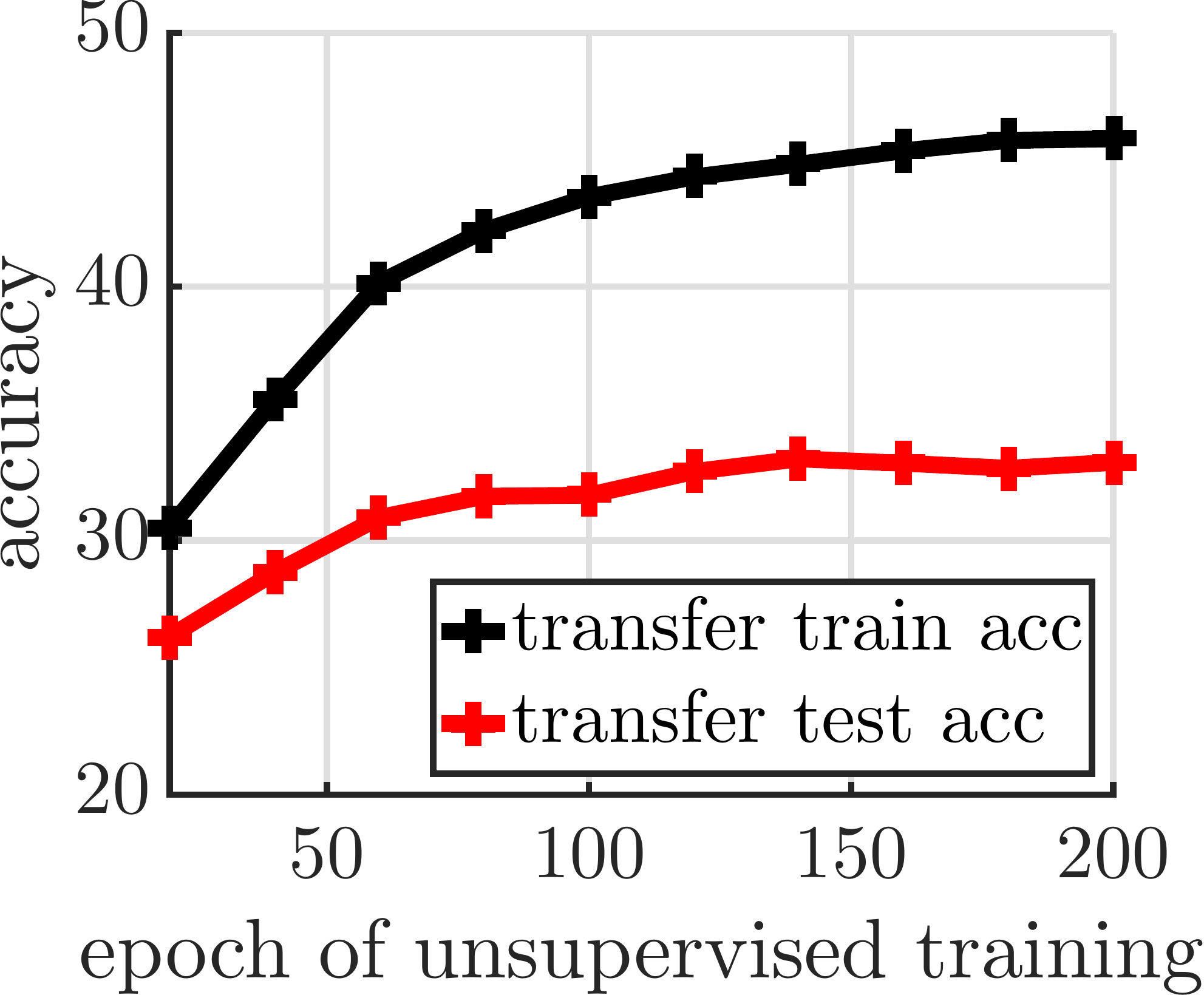}
  \includegraphics[height=10em]{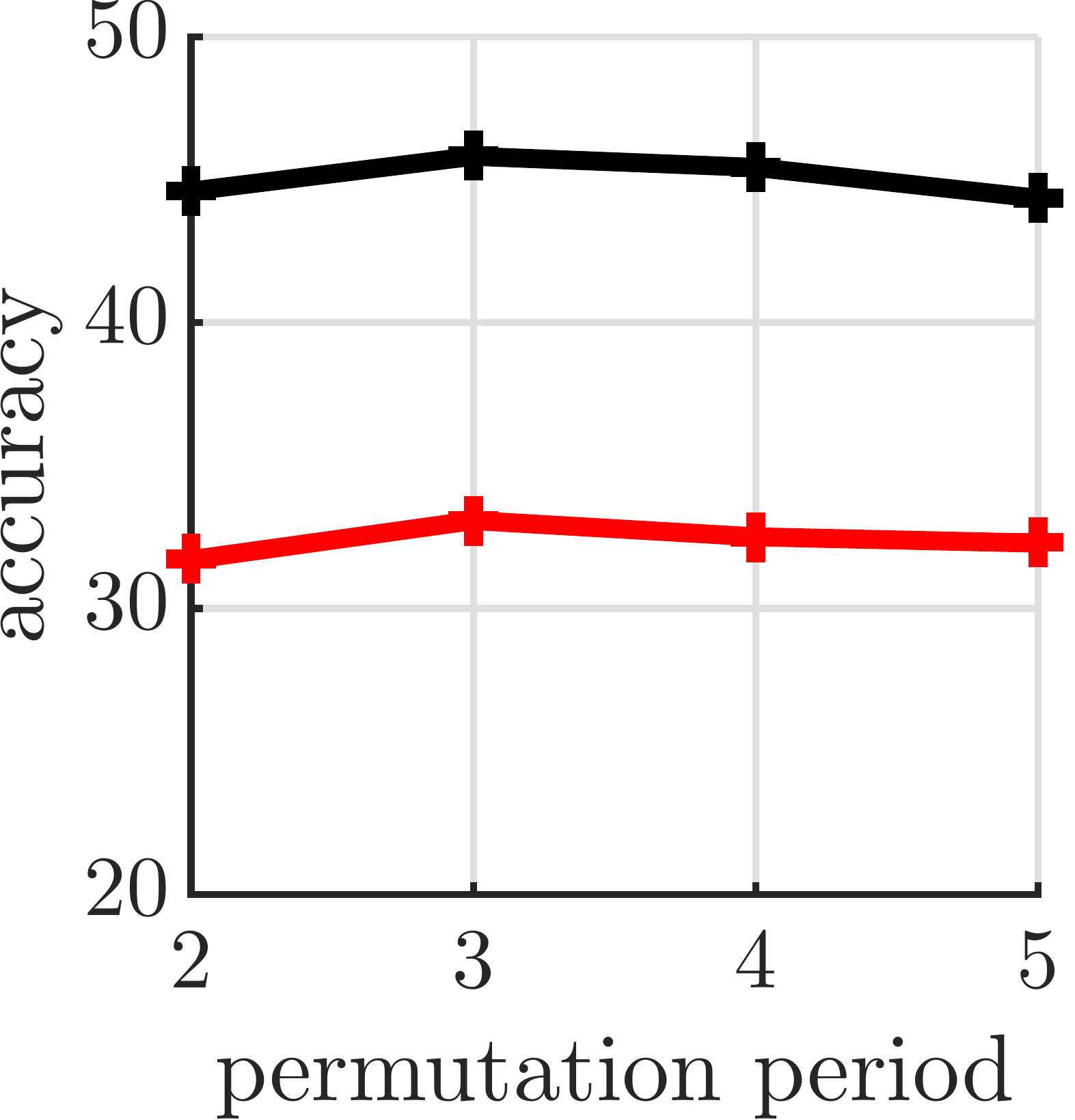}
  \caption{
    On the left, we measure the accuracy on ImageNet after training the features with different permutation rates
    There is a clear trade-off with an optimum at permutations performed every $3$ epochs.
    On the right, we measure the accuracy on ImageNet after training the features with our unsupervised approach as a function of the number of epochs.
    The performance improves with longer unsupervised training.
  }
  \label{fig:epochs}
\end{figure}

\paragraph{Evolution of the features.}
In this experiment, we are interested in understanding how the quality of our features evolves with the optimization of our cost function.
During the unsupervised training, we freeze the network every 20 epochs and learn a MLP classifier on top. We report the accuracy on the validation set
of ImageNet.
Figure~\ref{fig:epochs} shows the evolution of the performance on this transfer task as we optimize for our unsupervised approach.
The training performance improves monotonically with the epochs of the unsupervised training.
This suggests that optimizing our objective function correlates with learning transferable features, \ie, our features do not destroy useful
class-level information.
On the other hand, the test accuracy seems to saturate after a hundred epochs. This suggests that the MLP is overfitting rapidly on pre-trained features.

\paragraph{Effect of permutations.}
Assigning images to their target representations is a crucial feature of our approach.
In this experiment, we are interested in understanding how frequently we should update this assignment.
Indeed, updating the assignment, even partially, is relatively costly and may not be required to achieve
good performance.
Figure~\ref{fig:epochs} shows the transfer accuracies on ImageNet  as a function of the frequency of these updates.
The model is quite robust to choice of frequency, with a test accuracy always above $30\%$.
Interestingly, the accuracy actually degrades slightly with high frequency.
A possible explanation is that the network overfits rapidly to its own output, leading to relatively worse features.
In practice, we observe that updating the assignment matrix every $3$ epochs offers a good trade-off between
performance and accuracy.

\begin{figure*}[t!]
  \centering
  \begin{tabular}{ccccccc}
    \includegraphics[width=.11\linewidth]{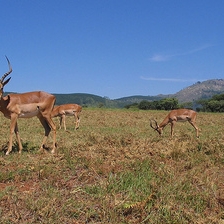} &
    \includegraphics[width=.11\linewidth]{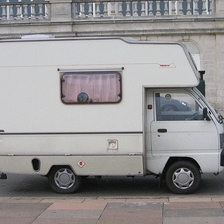} &
    \includegraphics[width=.11\linewidth]{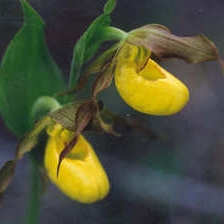} &
    \includegraphics[width=.11\linewidth]{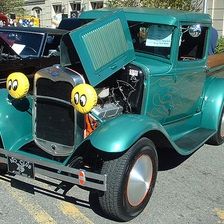} &
    \includegraphics[width=.11\linewidth]{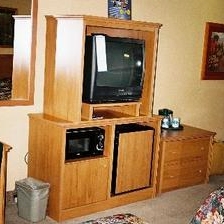} &
    \includegraphics[width=.11\linewidth]{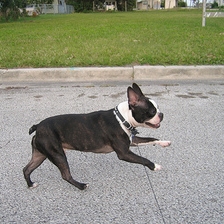} &
    \includegraphics[width=.11\linewidth]{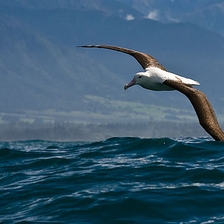} \\
    \includegraphics[width=.11\linewidth]{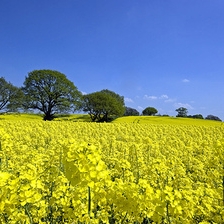}&
    \includegraphics[width=.11\linewidth]{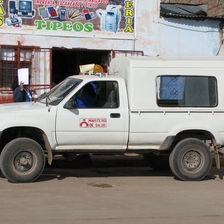} &
    \includegraphics[width=.11\linewidth]{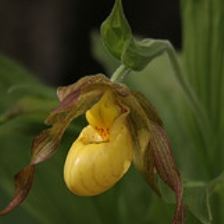} &
    \includegraphics[width=.11\linewidth]{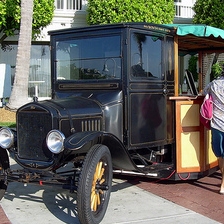} &
    \includegraphics[width=.11\linewidth]{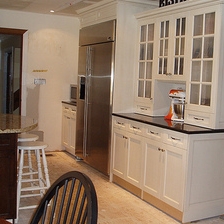} &
    \includegraphics[width=.11\linewidth]{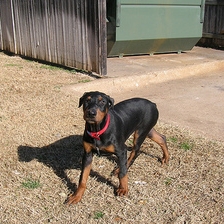} &
    \includegraphics[width=.11\linewidth]{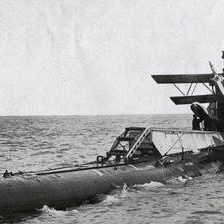} \\
    \includegraphics[width=.11\linewidth]{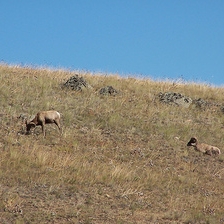} &
    \includegraphics[width=.11\linewidth]{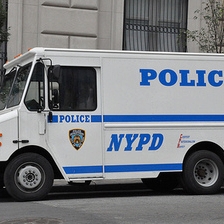} &
    \includegraphics[width=.11\linewidth]{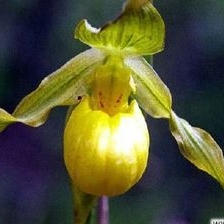} &
    \includegraphics[width=.11\linewidth]{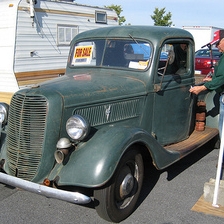} &
    \includegraphics[width=.11\linewidth]{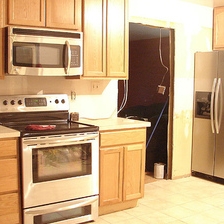} &
    \includegraphics[width=.11\linewidth]{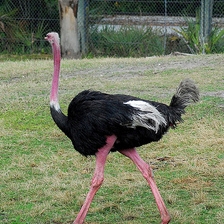} &
    \includegraphics[width=.11\linewidth]{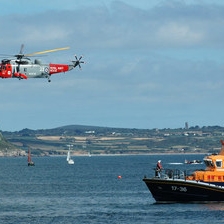}\\
    \includegraphics[width=.11\linewidth]{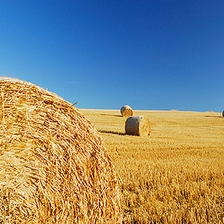}&
    \includegraphics[width=.11\linewidth]{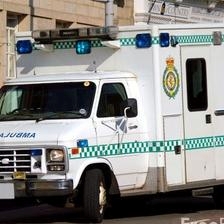} &
    \includegraphics[width=.11\linewidth]{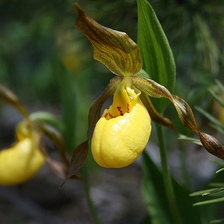} &
    \includegraphics[width=.11\linewidth]{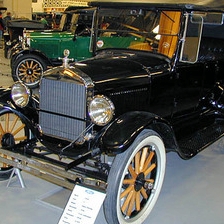} &
    \includegraphics[width=.11\linewidth]{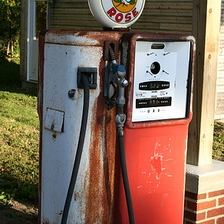} &
    \includegraphics[width=.11\linewidth]{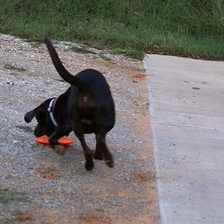} &
    \includegraphics[width=.11\linewidth]{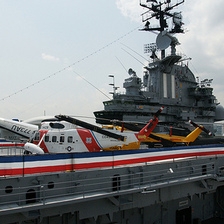}
  \end{tabular}
  \caption{
    Images and their $3$ nearest neighbors in ImageNet according to our model using an $\ell_2$ distance.
    The query images are shown on the top row, and the nearest neighbors are sorted from the closer to
    the further.
    Our features seem to capture global distinctive structures.
  }
  \label{fig:nn}
\end{figure*}

\begin{figure}[t]
  \centering
  \includegraphics[width=.49\linewidth]{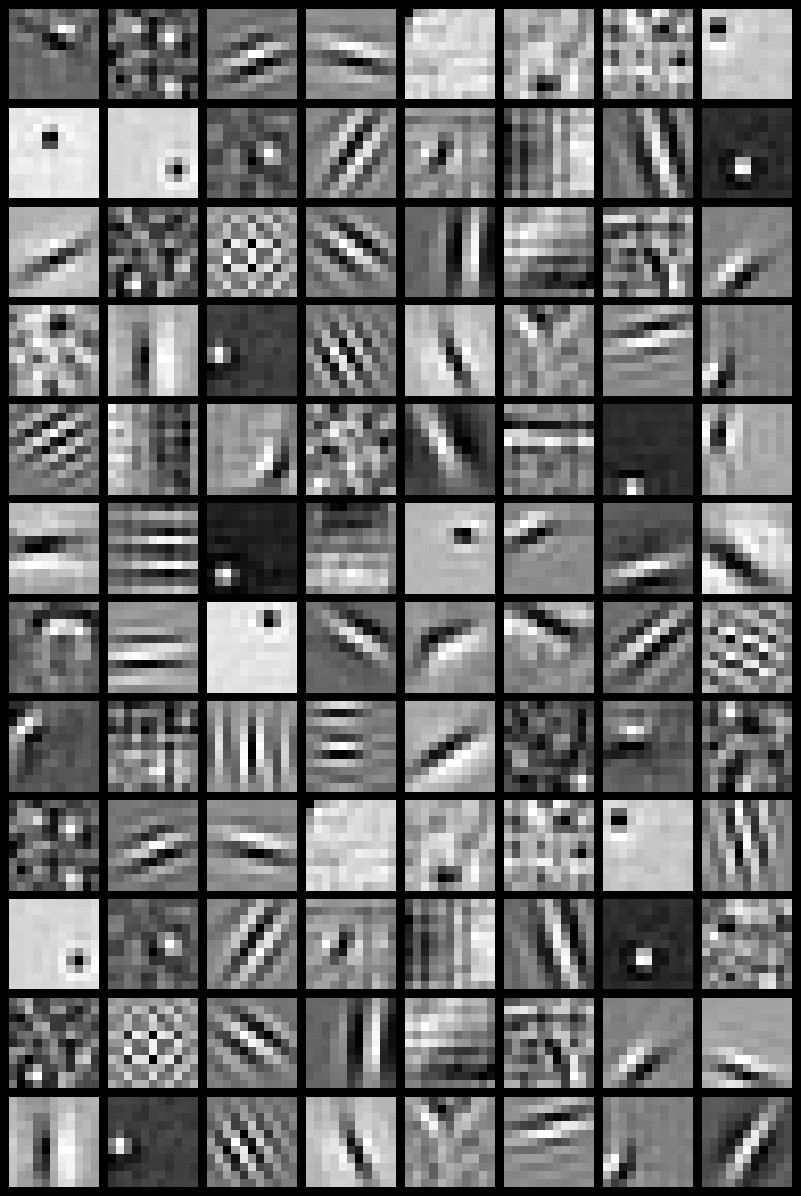}
  \includegraphics[width=.49\linewidth]{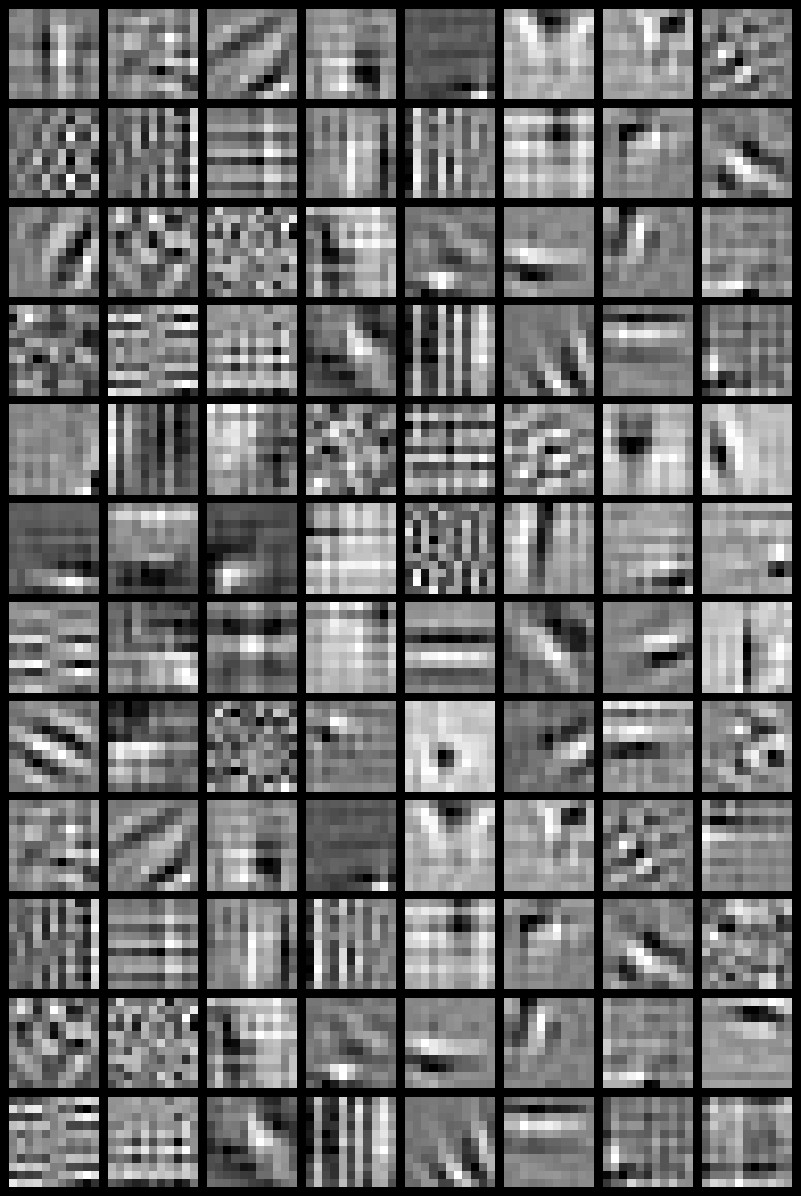}
  \caption{
    Filters form the first layer of an AlexNet trained on ImageNet with supervision (left) or with NAT (right).
    The filters are in grayscale, since we use grayscale gradient images as input.
    This visualization shows the composition of the gradients with the first layer.
  }
  \label{fig:filters}
\end{figure}

\paragraph{Visualizing the filters.}
Figure~\ref{fig:filters} shows a comparison between the first convolutional layer of an AlexNet trained with and without supervision.
Both take grayscale gradient images as input.
The visualization are obtained by composing the Sobel filtering with the filters of the first layer of the AlexNet.
Unsupervised filters are slightly less sharp than their supervised counterpart, but still maintain edge and orientation information.

\paragraph{Nearest neighbor queries.}
Our loss optimizes a distance between features and fixed vectors. This means that looking at the distance between
features should provide some information about the type of structure that our model captures.
Given a query image $x$, we compute its feature $f_\theta(x)$ and search for its nearest neighbors according to the $\ell_2$ distance.
Figure~\ref{fig:nn} shows images and their nearest neighbors.

The features capture relatively complex structures in images.
Objects with distinctive structures, like trunks or fruits, are well captured by our approach.
However, this information is not always related to true labels.
For example, the image of bird over the sea is matched to images capturing information about the sea or the sky rather than the bird.


\subsection{Comparison with the state of the art}
We report results on the transfer task both on ImageNet and \textsc{Pascal} VOC 2007.
In both cases, the model is trained on ImageNet.


\paragraph{ImageNet classification.}
In this experiment, we evaluate the quality of our features for the object classification task of ImageNet.
Note that in this setup, we build the unsupervised features on images that correspond to predefined image categories.
Even though we do not have access to category labels, the data itself is biased towards these classes.
In order to evaluate the features, we freeze the layers up to the last convolutional layer and train the classifier with supervision.
This experimental setting follows~\citet{NF16}.

We compare our model with several self-supervised approaches~\cite{WG15, DGE15, zhang2016colorful} and an unsupervised
approach, \ie,~\citet{DKD16}.
Note that self-supervised approaches use losses specifically designed for visual features.
Like BiGANs~\citep{DKD16}, NAT does not make any assumption about the domain but of the structure of its features.
Table~\ref{tab:in2in} compares NAT with these approaches.

\begin{table}[t]
  \centering
  \begin{tabular}{@{}lr@{}}
    \toprule
    Method& Acc@1 \\
    \midrule
    Random~\citep{NF16} & 12.0 \\
    \midrule
    SIFT+FV~\cite{akata2014good} & 55.6 \\
    \midrule
    \citet{WG15} & 29.8 \\
    \citet{DGE15} & 30.4 \\
    \citet{zhang2016colorful} & 35.2 \\
    $^1$\citet{NF16} & 38.1 \\
    \midrule
    BiGAN~\citep{DKD16} & 32.2 \\
    \midrule
    NAT & 36.0 \\
    \bottomrule
  \end{tabular}
  \caption{
    Comparison of the proposed approach to state-of-the-art unsupervised feature learning on ImageNet.
    A full multi-layer perceptron is retrained on top of the features.
    We compare to several self-supervised approaches and an unsupervised approach, \ie, BiGAN~\cite{DKD16}.
    $^1$\citet{NF16} uses a significantly larger amount of features than the original AlexNet.
    We report classification accuracy.
  }
  \label{tab:in2in}
\end{table}

Among unsupervised approaches, NAT compares favorably to BiGAN~\citep{DKD16}.
Interestingly, the performance of NAT are slightly better than \emph{self-supervised} methods, even though we do not
explicitly use domain-specific clues in images or videos to guide the learning.
While all the models provide performance in the $30-36\%$ range,
it is not clear if they all learn the same features.
Finally, all the unsupervised deep features are outperformed by hand-made features,
in particular Fisher Vectors with SIFT descriptors.
This baseline uses a slightly bigger MLP for the classifier and its performance can be improved by $2.2\%$
by bagging $8$ of these models.
This difference of $20\%$ in accuracy shows that unsupervised deep features are still quite far from
the state-of-the-arts among \emph{all} unsupervised features.


\paragraph{Transferring to \textsc{Pascal} VOC 2007.}
We carry out a second transfer experiment on the \textsc{Pascal} VOC dataset, on the classification and detection tasks.
The model is trained on ImageNet.
Depending on the task, we \emph{finetune} all layers in the network, or solely the classifier, following~\citet{DKD16}.
In all experiments, the parameters of the convolutional layers are initialized with the ones obtained with our unsupervised approach.
The parameters of the classification layers are initialized with gaussian weights.
We get rid of batch normalization layers and use a data-dependent rescaling of the parameters~\cite{krahenbuhl2015data}.
Table~\ref{tab:voc} shows the comparison between our model and other unsupervised approaches.
The results for other methods are taken from~\citet{DKD16} except for~\citet{zhang2016colorful}.

\begin{table}[t]
  \centering
  \begin{tabular}{@{}lccc@{}}
    \toprule
    & \multicolumn{2}{c}{Classification} & Detection \\
    \midrule
    Trained layers & fc6-8 & all & all \\
    \midrule
    ImageNet labels & 78.9 & 79.9 & 56.8 \\
    \midrule
    \citet{ACM15} & 31.0 & 54.2 & 43.9 \\
    \citet{pathak2016context} & 34.6 & 56.5 & 44.5\\
    \citet{WG15}  & 55.6 & 63.1 & 47.4 \\
    \citet{DGE15} & 55.1 & 65.3 & 51.1 \\
    \citet{zhang2016colorful} & 61.5 & 65.6 & 46.9 \\
    \midrule
    Autoencoder & 16.0 & 53.8 & 41.9 \\
    GAN & 40.5 & 56.4 & - \\
    BiGAN~\citep{DKD16} & 52.3 & 60.1 & 46.9 \\
    \midrule
    NAT & 56.7 & 65.3 & 49.4 \\
    \bottomrule
  \end{tabular}
  \caption{
    Comparison of the proposed approach to state-of-the-art unsupervised feature learning on VOC 2007 Classification and detection.
    We either fix the features after conv5 or we fine-tune the whole model.
    We compare to several self-supervised and an unsupervised approaches.
    The GAN and autoencoder baselines are from~\citet{DKD16}.
    We report mean average prevision as customary on \textsc{Pascal} VOC.
  }
  \label{tab:voc}
\end{table}

As with the ImageNet classification task, our performance is on par with self-supervised approaches, for both detection and classification.
Among purely unsupervised approaches, we outperform standard approaches like autoencoders or GANs by a large margin.
Our model also performs slightly better than the best performing BiGAN model~\citep{DKD16}.
These experiments confirm our findings from the ImageNet experiments.
Despite its simplicity, NAT learns feature that are as good as those obtained with more sophisticated and data-specific models.

\section{Conclusion}

This paper presents a simple unsupervised framework to learn discriminative features.
By aligning the output of a neural network to low-dimensional noise, we obtain
features on par with state-of-the-art unsupervised learning approaches.
Our approach explicitly aims at learning discriminative features, while most unsupervised approaches
target surrogate problems, like image denoising or image generation.
As opposed to self-supervised approaches, we make very few assumptions about the input space.
This makes our appproach very simple and fast to train.
Interestingly, it also shares some similarities with traditional clustering approaches as well as retrieval methods.
While we show the potential of our approach on visual data, it will be interesting to try other domains.
Finally, this work only considers simple noise distributions and alignment methods. A possible direction of research is to
explore target distributions and alignments that are more informative. This also would strengthen the relation
between NAT and methods based on distribution matching like the earth mover distance.

\paragraph{Acknowledgement.}
We greatly thank Herv\'e J\'egou for his help throughout the development of this project.
We also thank Allan Jabri, Edouard Grave, Iasonas Kokkinos, L\'eon Bottou, Matthijs Douze and the rest of FAIR
for their support and helpful discussion.
Finally, we thank Richard Zhang, Jeff Donahue and Florent Perronnin for their help.

\bibliographystyle{icml2016}
\bibliography{egbib}

\end{document}